# Deep learning-based detection of morphological features associated with hypoxia in H&E breast cancer whole slide images

Petru Manescu[1*], Joseph Geradts[2] and Delmiro Fernandez-Reyes[1]

1. Department of Computer Science, Faculty of Engineering Sciences, University College London, Gower Street, London, WC1E 6BT, United Kingdom.
2. Department of Pathology and Laboratory Medicine, East Carolina University Brody School of Medicine, Greenville, NC, USA.

**\*Corresponding Author:**
Dr. Petru Manescu (p.manescu@ucl.ac.uk). Department of Computer Science, Faculty of Engineering Sciences, University College London, Gower Street, London, WC1E 6BT, United Kingdom. Phone: +44 (0)7956 661869.

**Abstract**

Hypoxia occurs when tumour cells outgrow their blood supply, leading to regions of low oxygen levels within the tumour. Calculating hypoxia levels can be an important step in understanding the biology of tumours, their clinical progression and response to treatment. This study demonstrates a novel application of deep learning to evaluate hypoxia in the context of breast cancer histomorphology. More precisely, we show that Weakly Supervised Deep Learning (WSDL) models can accurately detect hypoxia associated features in routine Hematoxylin and Eosin (H&E) whole slide images (WSI). We trained and evaluated a deep Multiple Instance Learning model on tiles from WSI H&E tissue from breast cancer primary sites (n=240) obtaining on average an AUC of 0.87 on a left-out test set. We also showed significant differences between features of hypoxic and normoxic tissue regions as distinguished by the WSDL models. Such DL hypoxia H&E WSI detection models could potentially be extended to other tumour types and easily integrated into the pathology workflow without requiring additional costly assays.

**Introduction**

During the growth of malignant tumours, the neoplastic cells can outgrow their blood supply, leading to an insufficient delivery of oxygen and nutrients. As a result, regions within the tumour experience a shortage of oxygen creating an hypoxic[1] microenvironment which is a hallmark of solid tumours, including breast cancer[2,3]. Hypoxia can induce genetic instability in tumour cells, leading to the accumulation of genetic mutations and potentially driving tumour progression and resistance to treatment. For instance, hypoxic tumour cells have been shown to be more resistant to Radiation Therapy (RT) compared to well-oxygenated cells due to their decreased sensitivity to oxidative stress[4]. Adjuvant RT following breast-conserving surgery is part of standard treatment for early-stage breast cancer, as it considerably decreases the risk for ipsilateral breast tumour recurrence (IBTR)[2]. However, hypoxic regions within residual tumour foci may not receive an effective dose of radiation during treatment, leading to an increased the risk of recurrence[2,5]. Therefore, detecting hypoxia in solid primary tumours is important to identify potential patient-groups that do or do not benefit from certain therapies[4]. Moreover, hypoxia-induced changes in gene expression can facilitate the metastatic process, increasing the likelihood of cancer spreading to other parts of the body[1].

Hypoxia could be quantified based on the expression patterns of specific genes known to be regulated by hypoxia, such as those under the control of the hypoxia-inducible factor (HIF)-1. The expression levels of these Hypoxia-Regulated Genes (HRG) collectively indicate the degree of hypoxia in the tumour. Microarray or RNA sequencing technologies are usually employed to analyse the expression levels of such genes[6]. In the past years, a series of hypoxia-related prognostic signatures which attempt to correlate the HRG expressions to the clinical outcomes of breast cancer patient have been proposed[3,5,7]. Alternatively, hypoxia-related proteins, such as HIF-1α and its downstream targets, can be assessed using immunohistochemistry on tissue sections. The presence and localization of these proteins can indicate the extent of hypoxia in the tumour. Both approaches require separate assays and techniques, which may involve additional sample processing and costs outside the current breast cancer pathways and might be unavailable in some settings due to resource limitations. Haematoxylin and Eosin (H&E) stained tissue slides, on the other hand, are part of standard clinical practice and readily available in pathology laboratories. H&E slides provide rich morphological information, capturing cellular architecture, tissue organization, and tumour-stromal interactions. With the increasing adoption of Whole Slide Imaging (WSI) scanners, H&E slides routinely prepared for pathological examination are now being digitized and stored for Artificial Intelligence (AI) assisted analysis. Recent studies have shown the capabilities of Deep Learning (DL) models to predict gene expression, somatic mutations, or recurrence scores from H&E WSI of breast histology, showing potential to produce biologic insights and novel biomarkers[8,9].

Here we demonstrate a novel application of Weakly Supervised DL (WSDL) to detect hypoxia patterns in breast cancer tissue from H&E histology only. Contrary to fully supervised approaches, WS ones do not necessarily require tissue or cell level manual annotations. More precisely, we present a WSDL model (HypOxNet) that can evaluate hypoxia in the context of the tumour histomorphology, which was trained only with weak sample-level labels corresponding to HRG expression.

## RESULTS

The HypOxNet model was trained to classify WSI of H&E-stained breast tumor tissues (40x magnification) into hypoxic or normoxic (Fig. 1). We used data from The Cancer Genome Atlas (TCGA)[10] which contains both WSI of primary site tumours and a wide range of corresponding gene expressions measured by RNA sequencing (available at https://portal.gdc.cancer.gov/). The Hypoxia Hallmark signature from the Molecular Signature Database (MSigDB) collection[6] was used to stratify the tumour samples (n=240) into hypoxic and normoxic[3]. Details on this procedure can be found in[3]. This stratification of the samples represented the weak labels associated with each WSI.

**Deep learning accurately detects breast cancer hypoxia in H&E WSI**

We divided each WSI into small image patches (512x512 pixels) and discarded those containing less than 50% tissue (background tiles). We further employed these patches to train a Multiple Instance Learning (MIL) convolutional neural network model with attention pooling to predict sample-level labels (Fig. 1a). More specifically, the HypoOxNet model was trained to differentiate between "bags" of patch instances extracted from positive samples (hypoxic according to MSigDB) and "bags" of patch instances extracted from negative samples (normoxic). We trained and validated our WSDL HypOxNet model to predict hypoxia status with image patches of tissue from 120 normoxic samples and 120 hypoxic samples from the TCGA database[10]. The dataset was randomly split into training (2/3) and test (1/3) sets. A 3-fold cross validation yielded an average AUC of 0.868±0.038 (Fig 1b). Fig 1c shows the confusion matrix for one of the folds.

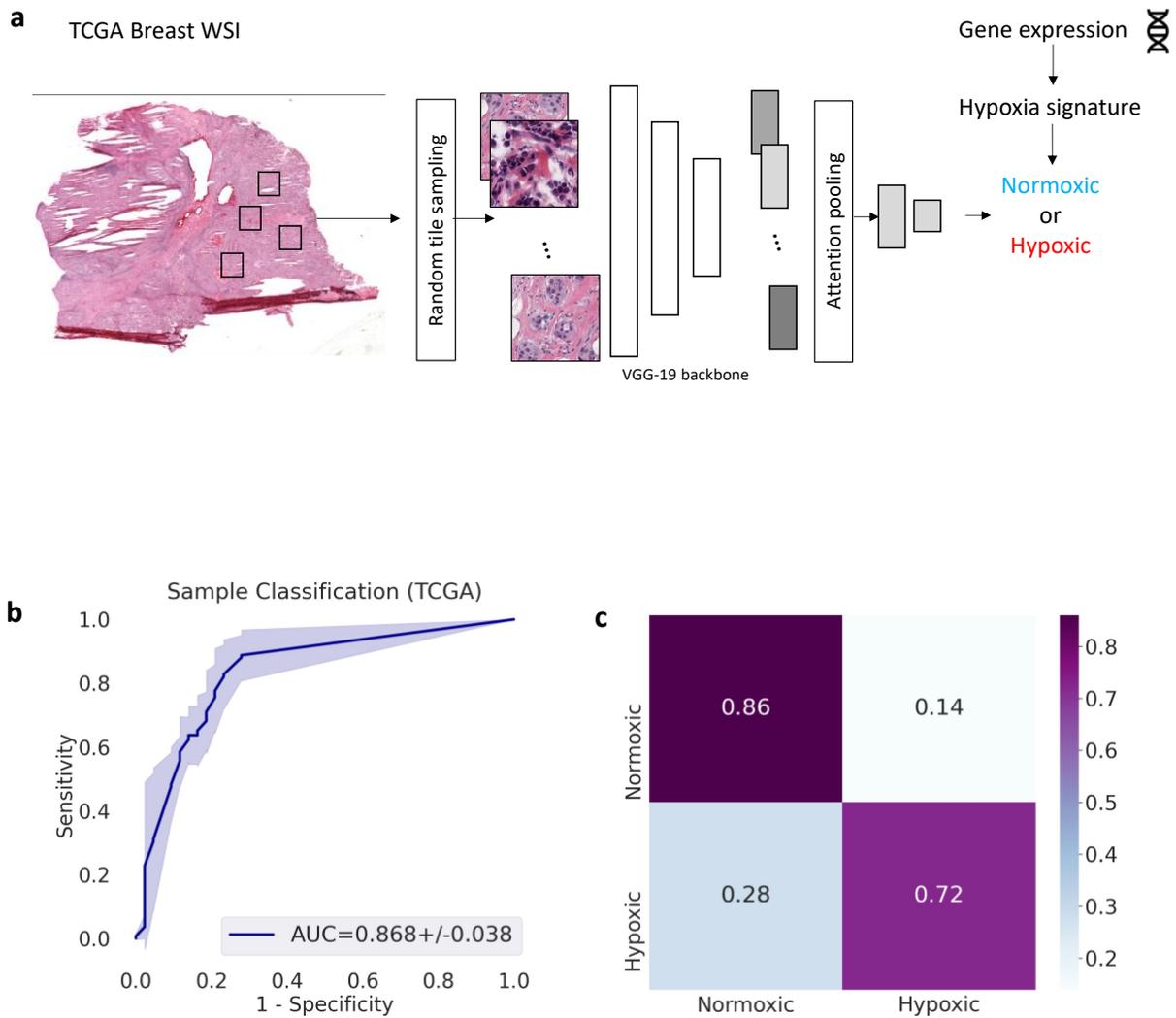

**Figure 1. Hypoxia prediction neural network (HypOxNet).** (a) Approach. Whole Slide Images of breast biopsies from the TCGA database are first divided into small tiles. Bags of tiles are passed through a Multiple Instance Learning ConvNet with an attention pooling[13] layer. The network is trained to classify the bags of tiles as hypoxic or normoxic, according to their hypoxia signature[3]. (b) Average AUROC on the left-out test set (n=80, 3 random train-test splits). (c) Confusion matrix on one of the test sets.

**Texture analysis reveals differences in hypoxic tissues**

We further looked at the model predictions in terms of histomorphology and image features. For a trained model, we randomly sampled 200 image patches (10 patches per sample) from 10 normoxic and 10 hypoxic samples from the test set. Next, we passed each single patch through the trained model ("bags" of one instance) and discarded those with scores lower than 0.9. Fig 2a-2b shows examples of tiles classified as "normoxic" by the model while Fig 2c-2d are examples of tiles classified as "hypoxic". Fig 2e-2f shows the tile regions highlighted as important for the classification (blue to red) by the Class Activation Maps (CAM).

For each of the patches classified as normoxic or hypoxic with high confidence (a score higher than 0.9) (n=156), we calculated the grey level co-occurrence matrix (GLCM) based texture features[14]. The homogeneity (Fig 2g), energy (Fig 2h) and correlation (Fig 2i) texture feature average values were significantly lower when calculated on the hypoxic tissue tiles than on those classified by the model as normoxic.

**Shape analysis reveals differences in macrophage cells from hypoxic tissue**

We next investigated the model predictions in terms of cell morphology. For this study, we used the tiles from breast tissues from the MoNuSaC challenge dataset[15] which contains patches with cell level contour annotations for various types of cells including macrophages. Similar to the previous experiment, we used a previously trained HypOxNet model to classify patches containing macrophage cells as normoxic or hypoxic and discarded those with scores lower than 0.9. Fig 3a shows an example of a tile containing macrophage cells classified as "normoxic" while Fig 3b is a tile classified as hypoxic. Fig 3c shows the "hypoxic" tile regions highlighted as important for the classification (blue to red) by the CAM. For each of the patches containing macrophages classified as normoxic or hypoxic with high confidence we used the corresponding manual annotations available in the dataset to compute the morphological descriptors of the macrophage cells (Fig 3d-3f). This preliminary analysis suggests that macrophage cells belonging to regions classified as hypoxic by the MIL model are not as "round" as macrophage cells in normoxic regions (circularity in Fig 3e). The analysis also showed differences in other shape metrics such as eccentricity (Fig 3d) and extent (Fig 3f).

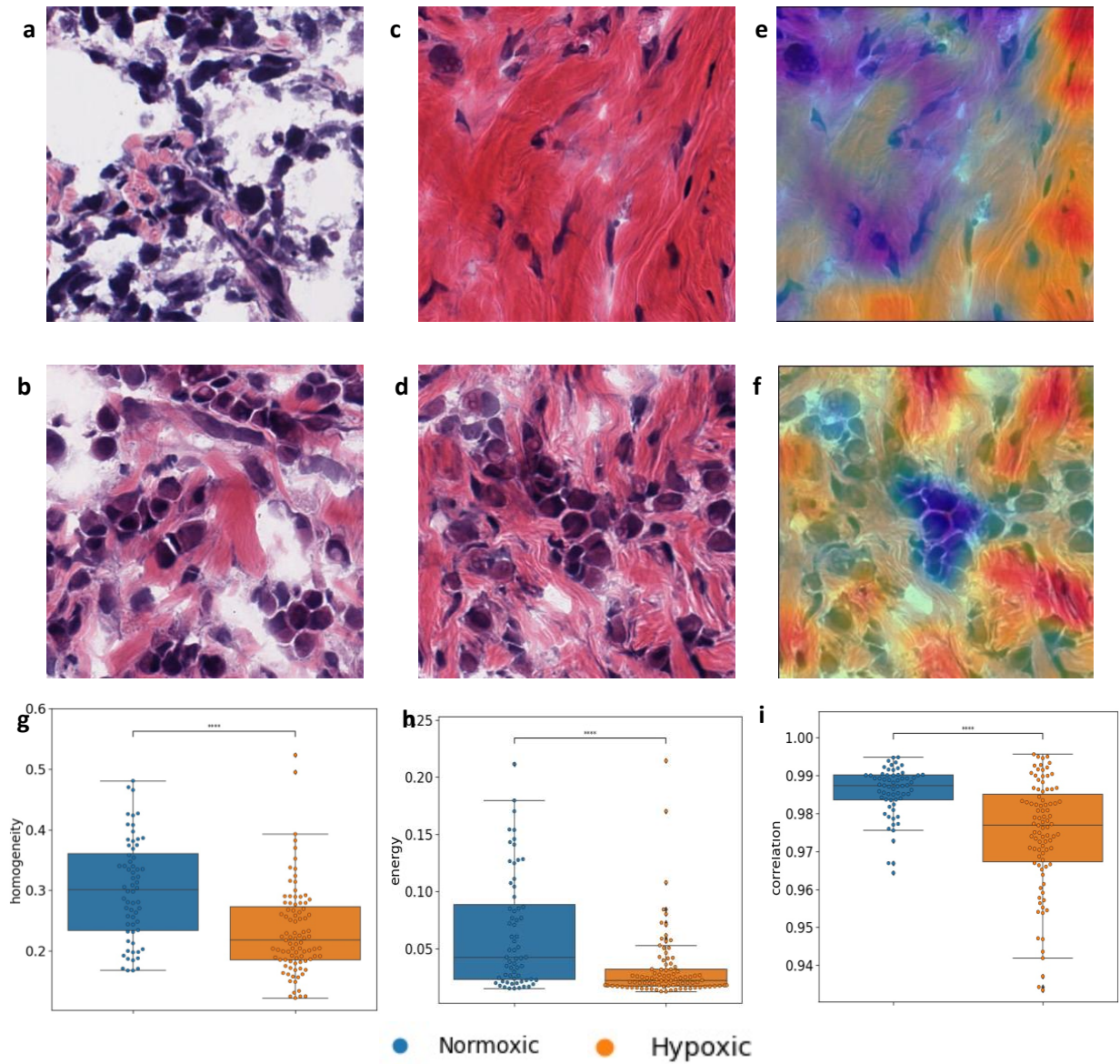

**Figure 2. Tile-level texture analysis.** (a)-(b). Examples of tiles classified by HypOxNet as normoxic. (c)-(d) Examples of tiles classified by HypOxNet as hypoxic. (e)-(f). Corresponding class activation maps (CAM) overlayed on the hypoxic tiles.

(g) Gray level co-occurrence matrix (GLCM) based texture features boxplots of individual tiles (n=156, size=512x512) classified by as hypoxic (orange) and normoxic (blue). *: p<0.05, **: p<0.01, ***: p<0.001, ****: p<0.0001.

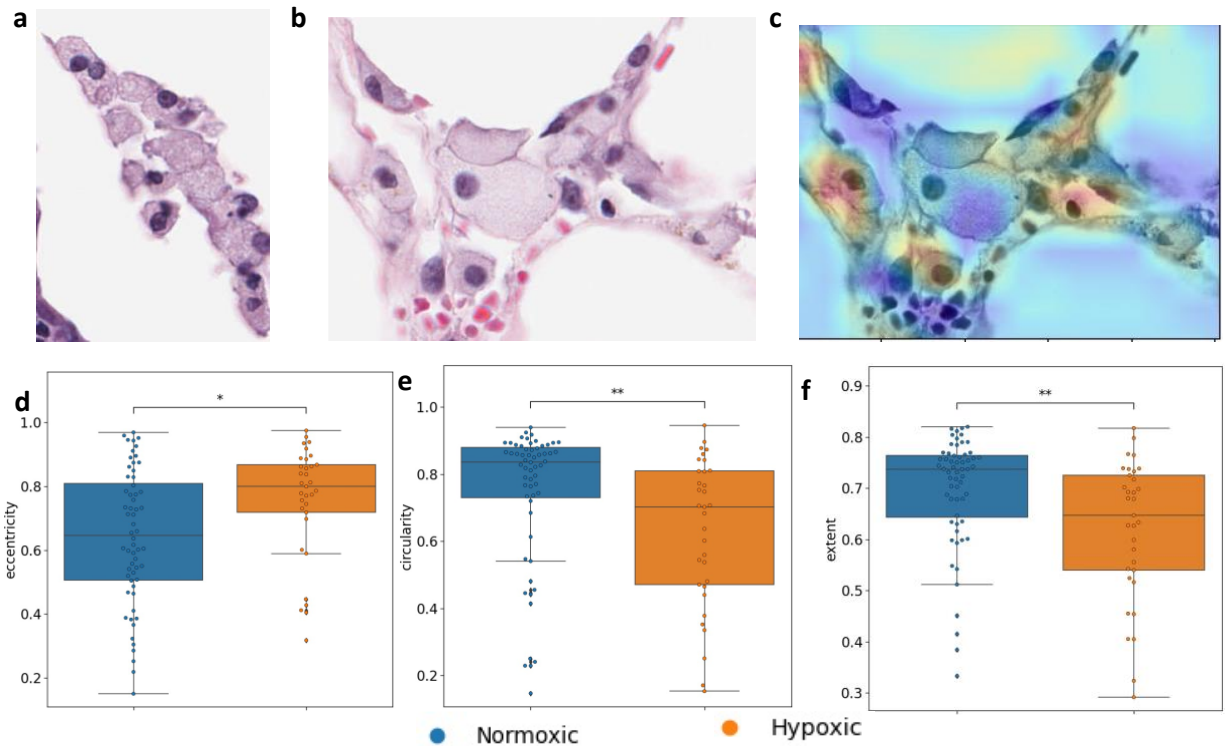

**Figure 3.** Cell level shape analysis of macrophages from the MoNuSaC annotated dataset[15] (n=96). (a). Example of tiles containing macrophages classified by HypOxNet as normoxic. (b) Example of tiles classified by HypOxNet as hypoxic. (c) Corresponding class activation maps (CAM) overlayed on the hypoxic tile. (d)-(e)-(f) Boxplots of selected binary shape descriptors of macrophage cells in tiles classified as normoxic (blue) and hypoxic (orange) by the DL model. *: p<0.05, **: p<0.01, ***: p<0.001, ****: p<0.0001. Shape descriptors were computed using the implementation available at: https://scikit-image.org/docs/stable/api/skimage.measure.html#skimage.measure.regionprops.

**DISCUSSION**

This study demonstrates that a WSDL model can distinguish hypoxia in the context of the tumour's histopathological characteristics from routine H&E slides. More precisely we show that H&E slides contain histomorphology biomarkers of hypoxia.

Gene expression profiling and hypoxia scores typically require separate assays and techniques, which may involve additional sample processing and costs. AI analysis of H&E slides, on the other hand, utilizes existing pathology slides, eliminating the need for additional experiments. H&E slides are part of standard clinical practice and readily available in pathology laboratories. Such a DL hypoxia detection model could potentially be easily integrated into the pathology workflow which would streamline the process and facilitate clinical adoption. Moreover, gene expression profiling or advanced molecular analyses may be less accessible in resource constrained settings. AI-based analysis of H&E slides could be a more feasible alternative in such situations. Linking AI-detected hypoxia patterns in H&E slides with clinical outcomes could potentially help establish prognostic or predictive significance, enabling faster patient stratification and treatment decision-making.

Our preliminary analysis suggests that macrophages in hypoxic regions have significant morphological differences compared to ones belonging to non-hypoxic regions. Tumour-associated macrophages (TAMs) play a crucial role in the tumour microenvironment and have a significant impact on tumour development. Hypoxia induces several changes in TAMs that promote tumour progression such as immunosuppression, angiogenesis or extracellular matrix remodelling[16,17].

As a proof-of-principle, we showed that HypOxNet can be successfully applied to breast cancer. However, this approach could potentially be applied to other tumour types as well, where hypoxia is an important factor in treatment selection[18–20]. More generally, our data provide evidence that weakly supervised deep learning models are suitable to predict molecular tumour phenotypes based on routine pathologic images.

## METHODS

For the experiments described above, we (1) tiled the regions containing tissue from each WSI, (2) trained a multiple instance learning convolutional neural network model with sample-level labels and (3) applied the trained model and reported the results.

**Network architecture and training details**

The convolutional layers of the HypOxNet model were initialized with weights from a VGG-19 model[11] pre-trained on the ImageNet dataset[12]. Tiles ($T_k^i$) with i=1,..,N corresponding to tissue regions were extracted from each sample *k* as p described in the previous section. The model was further trained to classify "bags" of these tiles with sample-level labels (*$L_k$*) provided by the hypoxia hallmark signature (MSigDB)[3]. The convolutional feature vectors corresponding to each tile ($F_k^i = conv(T_k^i)$) of the model were first reduced to 512 features each using Global Average Pooling (GAP) and then fused into a single "bag" level feature vector using Attention Pooling[13] (AP) followed by one dense layer and a classification layer:

$$L_k = \textbf{softmax}((\boldsymbol{ReLu}(AP(F_k^1, F_k^2, \ldots, F_k^N) \cdot W_1 + b_1)) \quad (1)$$

where *N* is the number of input image patches, $W_1$, $b_1$ are the corresponding weights of the dense layer. During training, twenty image tiles were randomly selected per sample for each iteration. These tiles were subject to on-the-fly geometrical augmentation (random rotations and random flips) as well as spectral augmentation (random hue modification, random gamma corrections, random noise). We employed stochastic gradient descent with a learning rate of 0.0003 and a cross entropy loss function to optimize the model weights during 100 epochs. At testing time, all the image tiles from each sample were passed through the trained models. The models were created using TensorFlow in Python 3.6.

**Morphology tile analysis**

For the texture analysis of tiles in Fig. 2, for each tile we converted the RGB tiles to gray-scale and extracted the average gray level co-occurrence matrix (GLCM) texture features implemented in Python's scikit-image (https://scikit-image.org/) toolkit as defined in[12]:

$$\text{Homogeneity} = \sum_i^N \sum_j^N \frac{P(i,j)}{1+(i-j)^2} \qquad (2)$$

$$\text{Energy} = \sqrt{\sum_i^N \sum_j^N P(i,j)^2} \qquad (3)$$

$$\text{Correlation} = \sum_i^N \sum_j^N \frac{(i-\mu_i)(j-\mu_j)}{\sqrt{\sigma_i^2 \sigma_j^2}} \qquad (4)$$

Where N denotes the number of grey levels (256), $P(i,j)$ is the grey-scale normalized value at position $i$ and $j$ of the patch.

Textural features were generated for each tile converted to grey-scale with distance at 1, rotation at 0, 45°, 90°, and 135° and then averaged (https://scikitimage.org/docs/0.7.0/api/skimage.feature.texture.html#skimage.feature.texture.greycomatrix).

For the morphology analysis of macrophages in Fig. 3, we extracted the classical binary shape descriptors available in the scikit-image toolkit (area, eccentricity, extent, diameter, perimeter, and solidity).

$$\text{Eccentricity} = \sqrt{1 - \frac{b^2}{a^2}} \qquad (5)$$

Where b = minor axis length, and a = major axis length of a connected component binary mask. An elongated object would have an eccentricity value close to 1, while a round object would have an eccentricity value close to 0.

$$\text{Circularity} = \frac{4\pi \cdot Area}{Perimeter^2} \qquad (6)$$

A perfect circle would have a circularity value of 1 while the value goes down as far as 0 for highly non-circular shapes.

$$\text{Extent} = \frac{A_{object}}{A_{BBox}} \qquad (7)$$

Where $A_{object}$ represents the area of the object (in pixels) while $A_{BBox}$ represents the area of its bounding box.

**Competing Interests**

The authors have declared that no competing interests exist.

**Data Availability Statement**

All the datasets used in this study are publicly available.

**Funding**

This work was supported by the UK National Institute for Health and Care Research (NIHR205968) and the UK Engineering and Physical Sciences Research Council (EP/P028608/1).

**Author Contributions**

PM, JG and DFR designed the study. PM did the experimental analysis. PM designed and developed the machine learning platform and its code. PM, DFR, JG analyzed the data. PM prepared the manuscript with contributions from all authors.

**SUPPLEMENTARY FIGURES**

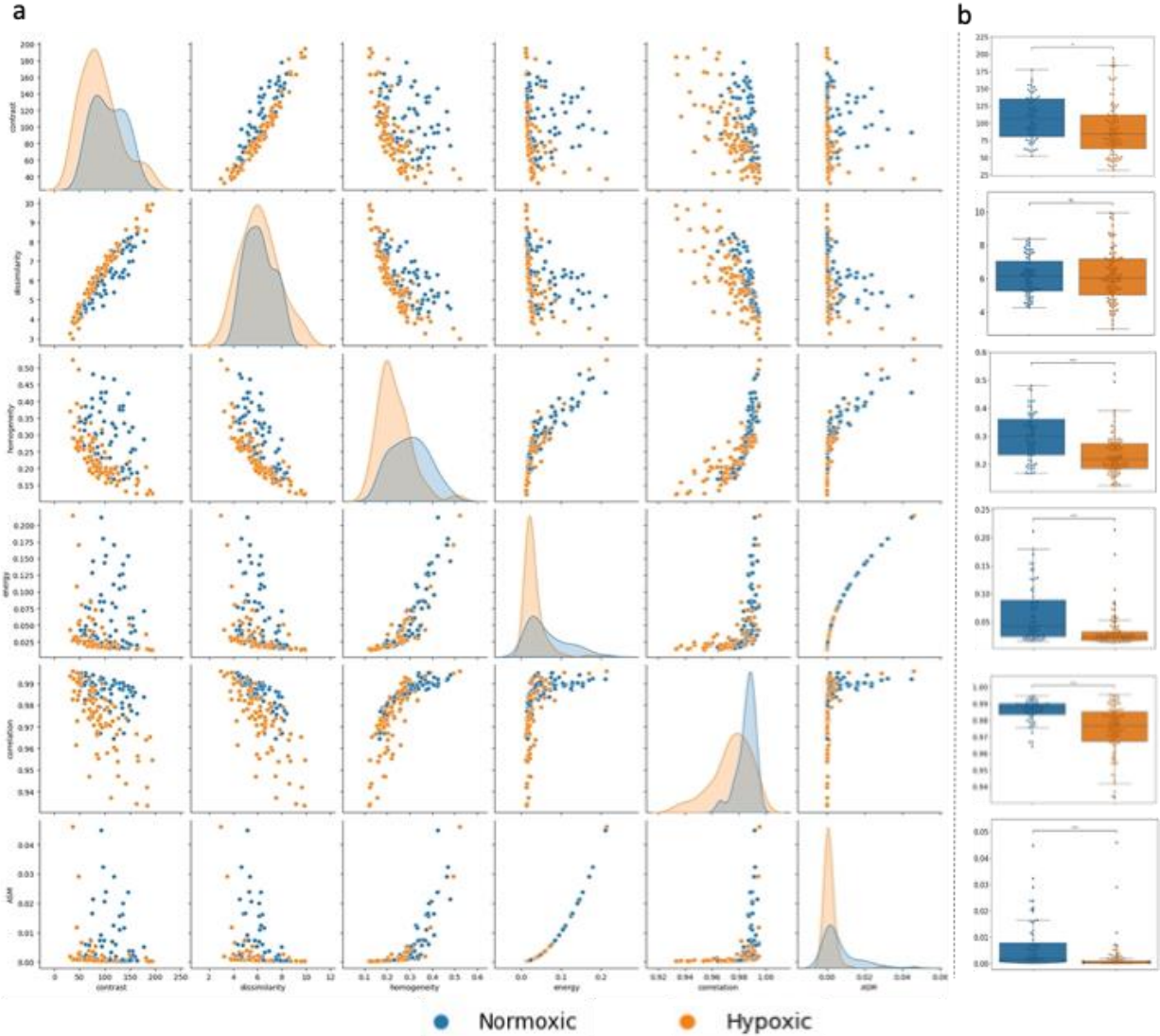

**Supplementary Figure 1**. Tile-level texture analysis (a) All gray level co-occurrence matrix (GLCM) based texture features pairplot of individual tiles (n=156, size=512x512) classified by HypOxNet as hypoxic and normoxic. (b) Corresponding boxplots.

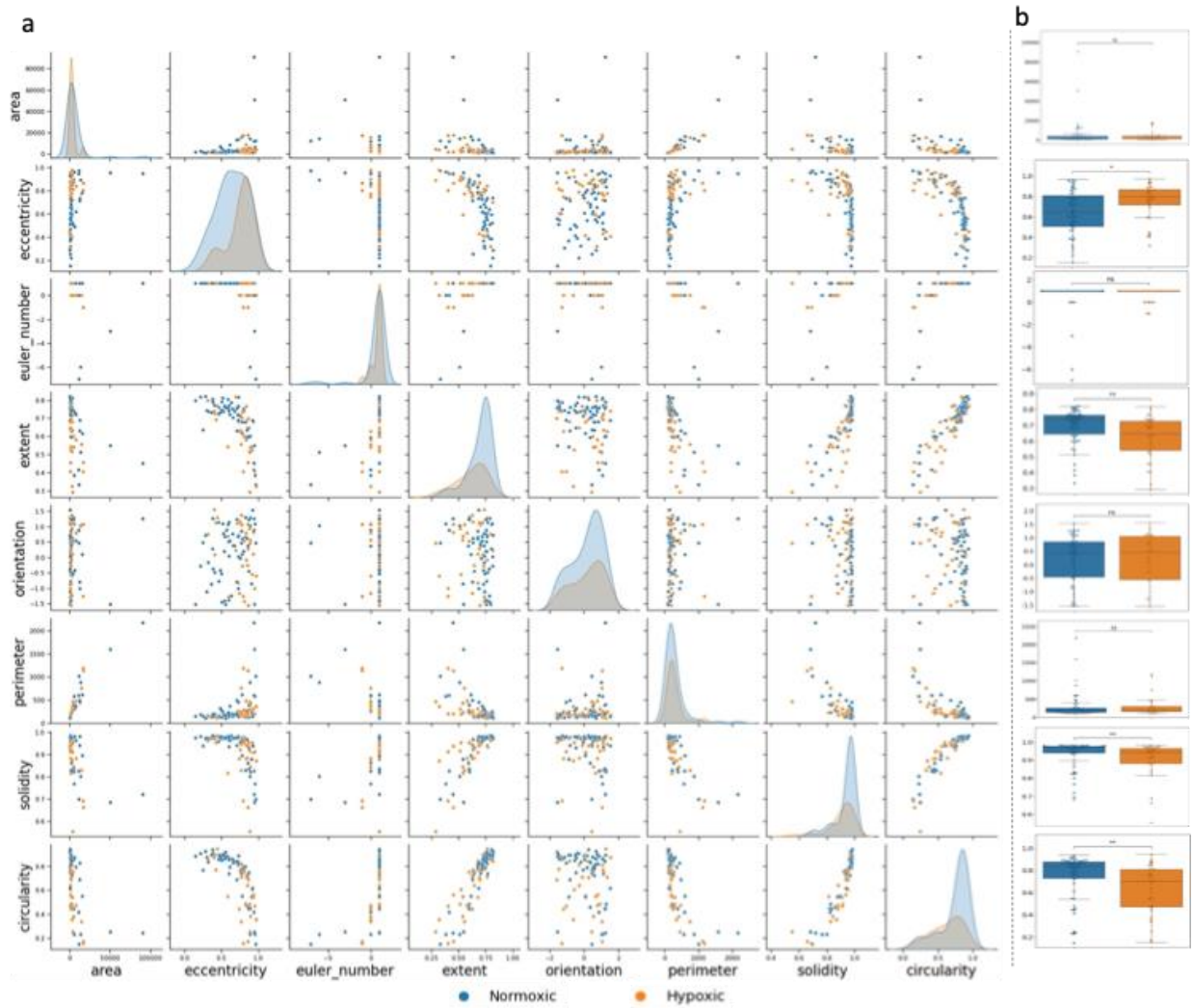

**Supplementary Figure 2.** Cell level shape analysis of macrophages from the MoNuSaC annotated dataset (n=96). (a) Pair plot of all binary shape descriptors of macrophage cells in tiles classified as normoxic and hypoxic by HypoNet. (b) Corresponding boxplots.